\documentclass[eat,twocolumn]{jmlr}

\usepackage{longtable}
\usepackage[load-configurations=version-1]{siunitx} 
\usepackage{caption}
\usepackage{sidecap}
\usepackage{multirow}
\usepackage[utf8]{inputenc} 
\usepackage[T1]{fontenc}    
\usepackage{url}
\usepackage{longtable}
\usepackage{booktabs}
\usepackage[load-configurations=version-1]{siunitx}
\usepackage{verbatim}
\usepackage{lscape}
\usepackage{rotating}

\jmlrvolume{}
\firstpageno{1}

\jmlryear{2022}
\jmlrworkshop{Machine Learning for Health (ML4H) 2022}

\title[Time-Varying Risks with COVID-19]{Estimating Discontinuous Time-Varying Risk Factors and Treatment Benefits for COVID-19 with Interpretable ML}

\author{\Name{Benjamin J. Lengerich}\Email{blengeri@mit.edu}\\
\addr Massachusetts Institute of Technology
\AND
\Name{Mark E. Nunnally}\Email{mark.nunnally@nyulangone.org}\\
\addr NYU Langone Health
\AND
\Name{Yin Aphinyanaphongs}\Email{yaphinyanaphongs@nyulangone.org}\\
\addr NYU Langone Health
\AND
\Name{Rich Caruana}\Email{rcaruana@microsoft.com}\\
\addr Microsoft Research
}

\begin{document}
\maketitle

\begin{abstract}
Treatment protocols, disease understanding, and viral characteristics changed over the course of the COVID-19 pandemic; as a result, the risks associated with patient comorbidities and biomarkers also changed. 
We add to the conversation regarding inflammation, hemostasis and vascular function in COVID-19 by performing a time-varying observational analysis of over 4000 patients hospitalized for COVID-19 in a New York City hospital system from March 2020 to August 2021. 
To perform this analysis, we apply tree-based generalized additive models with temporal interactions which recover discontinuous risk changes caused by discrete protocols changes. 
We find that the biomarkers of thrombosis increasingly predicted mortality from March 2020 to August 2021, while the association between biomarkers of inflammation and thrombosis weakened. 
Beyond COVID-19, this presents a straightforward methodology to estimate unknown and discontinuous time-varying effects.
\end{abstract}

\section{Introduction}
\label{sec:intro}
Treatment protocols, treatment availability, disease understanding, and viral characteristics have changed over the course of the COVID-19 pandemic; as a result, the risks associated with patient comorbidities and biomarkers have also changed. 
Analyses of hospitalized patients have identified that inflammatory biomarkers correspond to case severity \citep{liu2020neutrophil,garcia2020immune,lengerich2022automated}; in addition, observations of clotting have suggested that hemostasis and vascular function are critically dysregulated processes in patients hospitalized with COVID-19 \citep{tang2020abnormal,cui2020prevalence,kalafatis2021covid,alam2021hypercoagulability,al2020covid,mei2021role}. We add to this conversation by performing a time-varying observational analysis of over 4000 patients hospitalized for COVID-19 in the New York University Langone Health hospital system from March 2020 to August 2021 to elucidate the changing impacts of thrombosis, inflammation, and other risk factors on in-hospital mortality. 
We find that the association between mortality risk and thrombosis biomarkers increased over time, suggesting an opportunity for improved care by identifying and targeting therapies for patients with elevated thrombophilic propensity. 
This strengthening association between thrombosis and mortality contrasts against a weakening association between biomarkers of inflammation and mortality.

\section{Materials and Methods}
\paragraph{Dataset}
Our dataset consists of patients hospitalized in the NYU Langone Health system who had lab-confirmed cases of Covid-19. 
The highest density of patients were admitted in April 2020 (days 30-50 of the pandemic); this period also saw many high-risk patients (Figure~\ref{fig:cohort}).

For each patient, we observe demographics, comorbidities, outpatient medications, initial in-patient vitals, and initial in-patient lab tests (listed in ~\ref{sec:s1})). 
To streamline analysis, we binarize each of the 11 continuous-valued lab tests into a discrete biomarker rule which approximately separates high-risk and low-risk regions (Table~\ref{tab:table1}). 
This binarization was accomplished by fitting the model to the training dataset using continuous-valued lab tests to produce continuous-valued risk plots (Figure~\ref{fig:binarize}); to reduce the dimensionality of these covariates we select a data-driven threshold which separates the high-risk and low-risk regions.
Finally, for analysis, we group the 11 biomarkers into 3 groups: thrombosis risk (high D-Dimer, high hematocrit), inflammatory risk (high C-Reactive protein, high Neutrophil/Lymphocyte ratio, low serum albumin), and other biomarkers. 

\paragraph{Methods}
We use a generalized additive model (GAM) with interactions to predict in-hospital mortality from risk factors on hospital admission. 
The model (Figure~\ref{fig:arch}) estimates an effect of each risk factor and an interaction with time for each lab test. 
We use the tree-based GAMs 
\citep{nori2019interpretml} which are invariant to all monotonic feature transforms and recover discontinuities in clinical practice \citep{lengerich2022death}.

\section{Results}
The risk model is high accuracy, achieving an ROC of $0.939 \pm 0.001$, outperforming a logistic regression model which achieves an ROC of $0.859 \pm 0.001$. 

\begin{figure*}[htbp]
\floatconts
  {fig:figure1}
  {\caption{{\bf (A)} The associations between biomarkers and mortality have changed over time. Biomarkers of inflammation risk (elevated C-reactive protein, low albumin, high Neutrophil/Lymphocyte ratio) were initially powerful predictors of in-hospital mortality, but have become less predictive over time. 
  In contrast, biomarkers of thrombosis risk (elevated D-Dimer, elevated hematocrit) are more predictive of mortality in August 2021 than during March 2020. This suggests that the successful treatment of  patients hospitalized with indicators of thrombosis risk lagged behind the treatment of other groups. {\bf (B)} The in-hospital mortality rate decreased over time for all patients, but at a reduced rate for patients satisfying at least one biomarker rule for thrombosis risk. {\bf (C)} Treatment protocols changed over time, with a trend toward glucocorticoid and anticoagulant (overwhelmingly prophylactic heparin) prescriptions for the majority of patients. We mark the dates of several important publications and the rise of the Delta strain in NYC along the horizontal axis.
  }}
  {\includegraphics[width=0.7\linewidth]{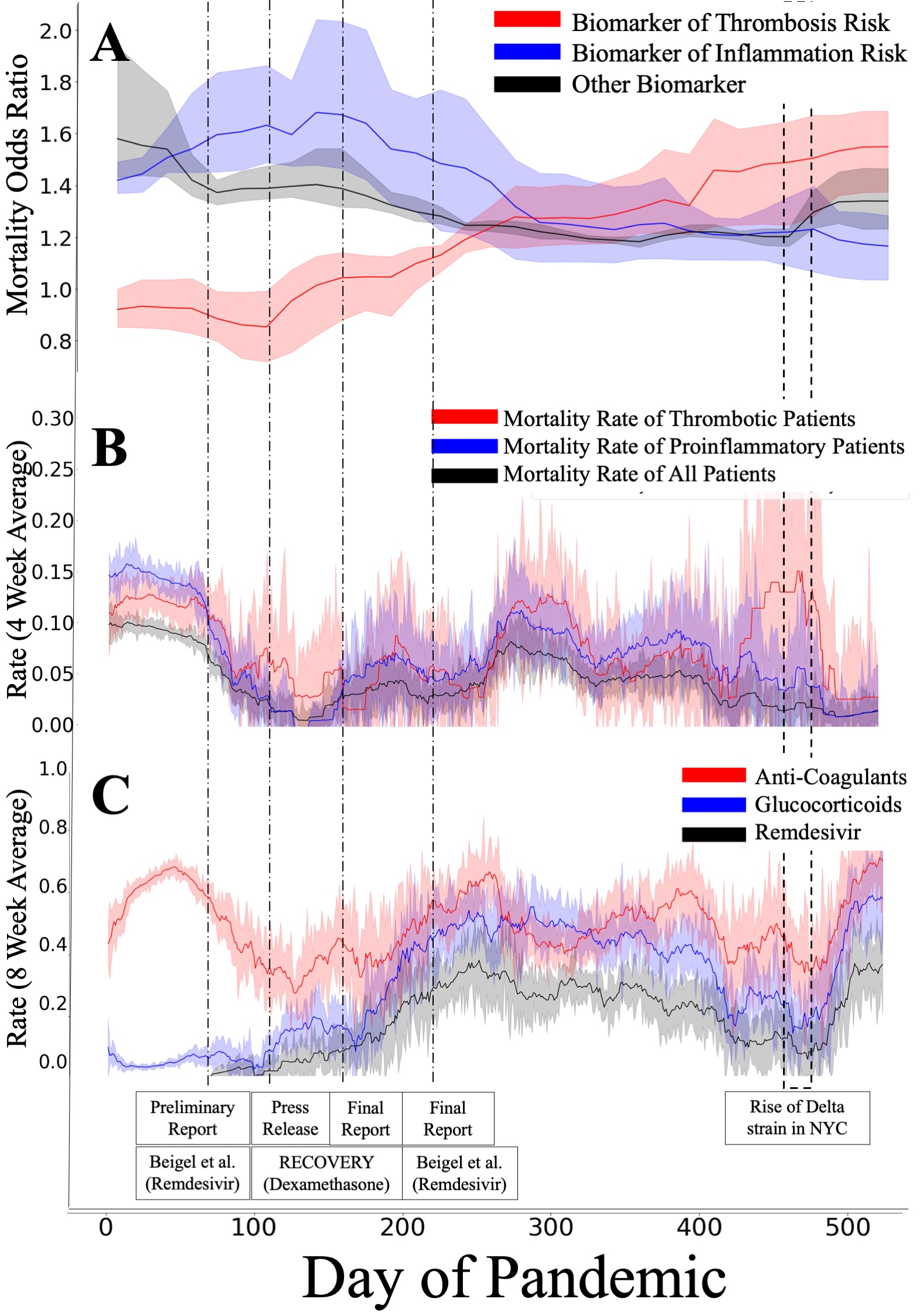}}
\end{figure*}

\begin{figure*}[ht]
    \floatconts
    {fig:all_effects}
    {\caption{Time-varying mortality risk associated with all biomarkers. Shaded regions indicate 95\% confidence intervals. These effects are estimated by the GAM after correcting for all other confounding effects underlying patient risk.}}
    {\includegraphics[width=\textwidth]{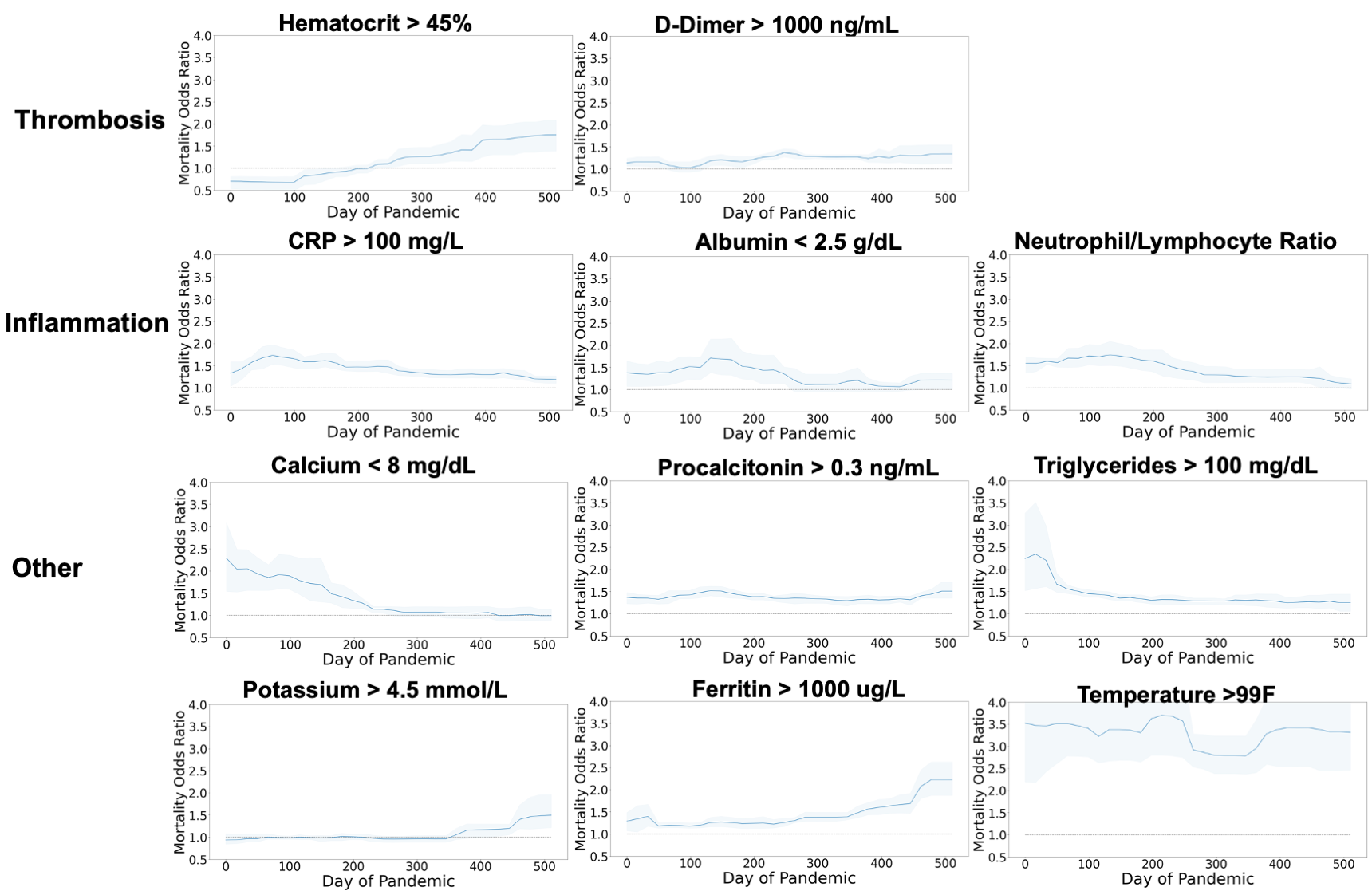}}
\end{figure*}

\begin{sidewaystable*}[hbtp]
\floatconts
  {tab:table1}
  {\caption{Biomarkers and rules analyzed by the GAM presented in Figure 1. In addition, we supplement the statistically more powerful GAM results with odds ratios of in-hospital mortality under (1) a univariable analysis without any correction for confounding factors, (2) a multivariable logistic regression model trained on patients admitted in the first 100 days of the pandemic, (3) a multivariable logistic regression model trained on patients admitted from days 100 to 300 of the pandemic, and (4) a multivariable logistic regression model trained on patients admitted after day 300 of the pandemic. The strongest risk factor is elevated temperature, and the only risk factors estimated to consistently increase in predictive power under the logistic regression model are elevated ferritin and elevated hematocrit.}}
  {\begin{tabular}{ccccccc}
  \toprule
  \bfseries Group & \bfseries Biomarker & \bfseries High-Risk Rule & 
\multicolumn{4}{c}{\bfseries Mortality Odds Ratio (95\% CI)} \\
  \midrule
  & & & \bfseries Univariable & \multicolumn{3}{c}{\bfseries Logistic Regression} \\ \midrule
  & & & & \footnotesize Day$<$100 (n=2827) & \footnotesize $100\leq$Day$<$300 (n=612) & \footnotesize Day$\geq$300 (n=821) \\ \midrule
  Thrombosis & D-Dimer & $>1000$ ng/mL & \footnotesize 2.24 (1.98, 2.58) & \footnotesize 1.21 (1.04, 1.47) & \footnotesize 0.93 (0.55, 1.65) & \footnotesize 1.23 (0.81, 1.91) \\
  Thrombosis & Hematocrit & $> 45\%$ & \footnotesize 1.05 (0.83, 1.23) & \footnotesize 0.89 (0.68, 1.35) & \footnotesize 1.90 (1.07, 2.97) & \footnotesize 1.61 (1.25, 1.94) \\
  Inflammation & CRP & $>100 mg/L$ & \footnotesize 3.04 (2.73, 3.42) & \footnotesize 1.94 (1.61, 2.39) & \footnotesize 1.54 (0.94, 2.07) & \footnotesize 1.66 (1.00, 3.23) \\
  Inflammation & NLR & $>7$ & \footnotesize 2.99 (2.71, 3.26) & \footnotesize 1.76 (1.44, 2.03) & \footnotesize 1.79 (1.10, 2.65) & \footnotesize 1.31 (0.82, 2.00) \\
  Inflammation & Albumin & $<2.5$ g/dL & \footnotesize 1.88 (1.61, 2.18) & \footnotesize 0.94 (0.70, 1.28)  & \footnotesize 1.55 (1.07, 2.12) & \footnotesize 1.25 (0.79, 1.79) \\
  Other & Temperature & $>99 F$ & \footnotesize 8.35 (7.29, 9.66) & \footnotesize 11.81 (7.23, 16.79) & \footnotesize 2.34 (1.29, 3.79) & \footnotesize 5.19 (3.28, 10.72) \\
  Other & Potassium & $>4.5$ mmol/L & \footnotesize 1.27 (1.13, 1.44) & \footnotesize 0.94 (0.79, 1.14) & \footnotesize 1.29 (0.85, 2.22) & \footnotesize 1.05 (0.81, 1.42) \\
  Other & Ferritin & $>1000$ ug/L & \footnotesize 2.20 (1.97, 2.46) & \footnotesize 1.29 (1.01, 1.56) & \footnotesize 2.00 (1.00, 2.72) & \footnotesize 2.18 (1.40, 3.42) \\
  Other & Calcium & $<8$ mg/dL & \footnotesize 2.28 (1.90, 2.63) & \footnotesize 1.56 (1.20, 2.12) & \footnotesize 0.89 (0.59, 1.28) & \footnotesize 1.66 (1.28, 2.47) \\
  Other & Triglycerides & $>100$ mg/dL & \footnotesize 2.15 (1.91, 2.39) & \footnotesize 1.78 (1.38, 2.31) & \footnotesize 1.62 (1.19, 2.76) & \footnotesize 1.59 (1.11, 2.12) \\
  Other & Procalcitonin & $>0.3$ ng/mL & \footnotesize 3.42 (3.07, 3.77) & \footnotesize 1.50 (1.17, 1.84) & \footnotesize 1.09 (0.85, 1.39) & \footnotesize 1.88 (1.43, 2.91) \\
  \bottomrule
  \end{tabular}}
\end{sidewaystable*}

\paragraph{Predictive Power of Thrombosis Biomarkers Increased}
The model estimates the time-varying contribution to mortality risk from each biomarker, and the mean mortality risk contributed by each biomarker group (thrombosis, inflammation, or other) after correcting for confounding from all other risk factors. 
The association between thrombosis biomarkers and in-hospital mortality strengthened over time (Figure~\ref{fig:figure1}A), rising from an OR of 0.92 (95\% CI 0.85-1.00) in March 2020 to an OR of 1.55 (1.38-1.69) in August 2021. 
The biomarkers indicating thrombosis risk are: Hematocrit > 45\%, D-Dimer > 1000 (the effects of these individual biomarkers are displayed in Figure~\ref{fig:all_effects}, Table~\ref{tab:table1}).

\paragraph{Predictive Power of Biomarkers of Inflammation Decreased}
The rise in thrombosis risk contrasts against a decrease in risk associated with inflammation-related biomarkers, which dropped from an OR of 1.42 (1.37-1.49) in March 2020 to 1.16 (1.03-1.28) in August 2021 (Figure~\ref{fig:figure1}A). 
While other biomarkers had stable impacts on mortality risk (e.g. the risk associated with elevated temperature remained consistently strong), the risk associated with elevated ferritin increased from an OR of 1.29 (1.07-1.49) to 2.22 (1.82-2.69). 
In particular, the predictive power of Neutrophil / Lymphocyte Ratio (NLR), a measure of inflammation and Covid-19 severity \citep{lagunas2020neutrophil,jimeno2021prognostic}, decreased (Figure~\ref{fig:all_effects}). 

\paragraph{Corroboration}
These trends are qualitatively corroborated by raw mortality rates (Figure~\ref{fig:figure1}B), LR (Table~\ref{tab:table1}), and correspond to trends in prescription rates (Figure~\ref{fig:figure1}C): clinical trials suggested the utility of glucocorticoids \citep{group2020dexamethasone} and remdesivir \citep{beigel2020remdesivir}, and the prescription rates of these treatments increased following publication of these studies. However, despite recognition of the importance of thrombosis and clotting in COVID-19 \citep{tang2020abnormal,cui2020prevalence,kalafatis2021covid,alam2021hypercoagulability,al2020covid,mei2021role}, anticoagulant prescription rates varied as the effectiveness of thromboprophylaxis with heparin was questioned \citep{remap2021therapeutic,helms2020high}.

\section{Discussion}
These results suggest that success in care for patients at risk for thrombosis lagged behind the success in care for patients with inflammation. 
As with all observational analyses, this study has limitations and cannot identify a singular cause of the trends. 
Several hypotheses would be consistent with these data, including: (1) the SARS-CoV-2 Delta strain shifted the importance of intrinsic risk factors, (2) successful efforts for early detection and treatment of thrombosis risk \citep{ierardi2021early} widened the difference in mortality risk between a serologically defined prothrombotic state and active thromboses, (3) potential interactions between anti-inflammation treatments and haemostasis, vascular function, and thrombophilic propensity \citep{orsi2021glucocorticoid,johannesdottir2013use}, (4) a lack of effective thromboprophylaxis treatments in COVID-19: there is little evidence for thromboprophylaxis from heparin in COVID-19 patients \citep{remap2021therapeutic,helms2020high} which may be due to heparin’s reliance on endogenous Antithrombin (AT) \citep{bussey2004heparin} which can be reduced in COVID-19 patients \citep{gross2020covid} --- anticoagulants such as Argatroban \citep{arachchillage2020anticoagulation} or Bivalirudin \citep{smith2020bivalirudin} which do not rely on AT may exert more powerful thromboprophylaxis than heparin in COVID-19 patients, or (5) an alternate process linked to thrombosis risk factors but also potentially implicating other aspects of endothelial and vascular dysfunction. 

\section{Conclusions}
GAMs with interactions are effective at identifying both homogeneous (static) and time-varying risk factors. 
When applied to rapidly-changing medical situations such as a pandemic outbreak, the appropriate risk model must accommodate both discontinuous risk factors influenced by treatment thresholds and discontinuous shifts in risk factors arising from changes in viral characteristics and treatment protocols. 
Tree-based GAMs recover high-resolution pictures of risk factors and treatment benefits, and interactions with time recover discontinuous effects. 
This approach reveals that for COVID-19 patients in NYC, the assocation between mortality and pro-inflammatory biomarkers decreased from March 2020 to August 2021 while the assocation between mortality and pro-thrombotic biomarkers strengthened.

\acks{
Yin Aphinyanaphongs was partially supported by NIH 3UL1TR001445-05 and National Science Foundation award \#1928614.
}

\bibliography{covid}

\begin{thebibliography}{26}
\providecommand{\natexlab}[1]{#1}
\providecommand{\url}[1]{\texttt{#1}}
\expandafter\ifx\csname urlstyle\endcsname\relax
  \providecommand{\doi}[1]{doi: #1}\else
  \providecommand{\doi}{doi: \begingroup \urlstyle{rm}\Url}\fi

\bibitem[Al-Samkari et~al.(2020)Al-Samkari, Karp~Leaf, Dzik, Carlson, Fogerty,
  Waheed, Goodarzi, Bendapudi, Bornikova, Gupta, et~al.]{al2020covid}
Hanny Al-Samkari, Rebecca~S Karp~Leaf, Walter~H Dzik, Jonathan~CT Carlson,
  Annemarie~E Fogerty, Anem Waheed, Katayoon Goodarzi, Pavan~K Bendapudi,
  Larissa Bornikova, Shruti Gupta, et~al.
\newblock Covid-19 and coagulation: bleeding and thrombotic manifestations of
  sars-cov-2 infection.
\newblock \emph{Blood}, 136\penalty0 (4):\penalty0 489--500, 2020.

\bibitem[Alam(2021)]{alam2021hypercoagulability}
Walid Alam.
\newblock Hypercoagulability in covid-19: A review of the potential mechanisms
  underlying clotting disorders.
\newblock \emph{SAGE open medicine}, 9:\penalty0 20503121211002996, 2021.

\bibitem[Arachchillage et~al.(2020)Arachchillage, Remmington, Rosenberg, Xu,
  Passariello, Hall, Laffan, and Patel]{arachchillage2020anticoagulation}
Deepa~J Arachchillage, Christopher Remmington, Alex Rosenberg, Tina Xu,
  Maurizio Passariello, Donna Hall, A~Mike Laffan, and Brijesh~V Patel.
\newblock Anticoagulation with argatroban in patients with acute antithrombin
  deficiency in severe covid-19.
\newblock \emph{British journal of haematology}, 2020.

\bibitem[Beigel et~al.(2020)Beigel, Tomashek, Dodd, Mehta, Zingman, Kalil,
  Hohmann, Chu, Luetkemeyer, Kline, et~al.]{beigel2020remdesivir}
John~H Beigel, Kay~M Tomashek, Lori~E Dodd, Aneesh~K Mehta, Barry~S Zingman,
  Andre~C Kalil, Elizabeth Hohmann, Helen~Y Chu, Annie Luetkemeyer, Susan
  Kline, et~al.
\newblock Remdesivir for the treatment of covid-19—preliminary report.
\newblock \emph{New England Journal of Medicine}, 2020.

\bibitem[Bussey et~al.(2004)Bussey, Francis, and Group]{bussey2004heparin}
Henry Bussey, John~L Francis, and Heparin~Consensus Group.
\newblock Heparin overview and issues.
\newblock \emph{Pharmacotherapy: The Journal of Human Pharmacology and Drug
  Therapy}, 24\penalty0 (8P2):\penalty0 103S--107S, 2004.

\bibitem[Cui et~al.(2020)Cui, Chen, Li, Liu, and Wang]{cui2020prevalence}
Songping Cui, Shuo Chen, Xiunan Li, Shi Liu, and Feng Wang.
\newblock Prevalence of venous thromboembolism in patients with severe novel
  coronavirus pneumonia.
\newblock \emph{Journal of Thrombosis and Haemostasis}, 18\penalty0
  (6):\penalty0 1421--1424, 2020.

\bibitem[Garc{\'\i}a(2020)]{garcia2020immune}
Luis~F Garc{\'\i}a.
\newblock Immune response, inflammation, and the clinical spectrum of covid-19.
\newblock \emph{Frontiers in immunology}, 11:\penalty0 1441, 2020.

\bibitem[Gross et~al.(2020)Gross, Moerer, Weber, Huber, and
  Scheithauer]{gross2020covid}
Oliver Gross, Onnen Moerer, Manfred Weber, Tobias~B Huber, and Simone
  Scheithauer.
\newblock Covid-19-associated nephritis: early warning for disease severity and
  complications?
\newblock \emph{The Lancet}, 395\penalty0 (10236):\penalty0 e87--e88, 2020.

\bibitem[Group(2020)]{group2020dexamethasone}
The RECOVERY~Collaborative Group.
\newblock Dexamethasone in hospitalized patients with covid-19—preliminary
  report.
\newblock \emph{The New England journal of medicine}, 2020.

\bibitem[Helms et~al.(2020)Helms, Tacquard, Severac, Leonard-Lorant, Ohana,
  Delabranche, Merdji, Clere-Jehl, Schenck, Gandet, et~al.]{helms2020high}
Julie Helms, Charles Tacquard, Fran{\c{c}}ois Severac, Ian Leonard-Lorant,
  Micka{\"e}l Ohana, Xavier Delabranche, Hamid Merdji, Rapha{\"e}l Clere-Jehl,
  Malika Schenck, Florence~Fagot Gandet, et~al.
\newblock High risk of thrombosis in patients with severe sars-cov-2 infection:
  a multicenter prospective cohort study.
\newblock \emph{Intensive care medicine}, 46\penalty0 (6):\penalty0 1089--1098,
  2020.

\bibitem[Ierardi et~al.(2021)Ierardi, Coppola, Fusco, Stellato, Aliberti,
  Andrisani, Vespro, Arrichiello, Panigada, Monzani, et~al.]{ierardi2021early}
Anna~Maria Ierardi, Andrea Coppola, Stefano Fusco, Elvira Stellato, Stefano
  Aliberti, Maria~Carmela Andrisani, Valentina Vespro, Antonio Arrichiello,
  Mauro Panigada, Valter Monzani, et~al.
\newblock Early detection of deep vein thrombosis in patients with coronavirus
  disease 2019: who to screen and who not to with doppler ultrasound?
\newblock \emph{Journal of ultrasound}, 24\penalty0 (2):\penalty0 165--173,
  2021.

\bibitem[Jimeno et~al.(2021)Jimeno, Ventura, Castellano, Garc{\'\i}a-Adasme,
  Miranda, Touza, Lllana, and L{\'o}pez-Escobar]{jimeno2021prognostic}
Sara Jimeno, Paula~S Ventura, Jose~M Castellano, Salvador~I Garc{\'\i}a-Adasme,
  Mario Miranda, Paula Touza, Isabel Lllana, and Alejandro L{\'o}pez-Escobar.
\newblock Prognostic implications of neutrophil-lymphocyte ratio in covid-19.
\newblock \emph{European Journal of Clinical Investigation}, 51\penalty0
  (1):\penalty0 e13404, 2021.

\bibitem[Johannesdottir et~al.(2013)Johannesdottir, Horv{\'a}th-Puh{\'o},
  Dekkers, Cannegieter, J{\o}rgensen, Ehrenstein, Vandenbroucke, Pedersen, and
  S{\o}rensen]{johannesdottir2013use}
Sigrun~A Johannesdottir, Erzs{\'e}bet Horv{\'a}th-Puh{\'o}, Olaf~M Dekkers,
  Suzanne~C Cannegieter, Jens Otto~L J{\o}rgensen, Vera Ehrenstein, Jan~P
  Vandenbroucke, Lars Pedersen, and Henrik~Toft S{\o}rensen.
\newblock Use of glucocorticoids and risk of venous thromboembolism: a
  nationwide population-based case-control study.
\newblock \emph{JAMA internal medicine}, 173\penalty0 (9):\penalty0 743--752,
  2013.

\bibitem[Kalafatis(2021)]{kalafatis2021covid}
Michael Kalafatis.
\newblock Covid-19: A serious vascular disease with primary symptoms of a
  respiratory ailment, 2021.

\bibitem[Lagunas-Rangel(2020)]{lagunas2020neutrophil}
Francisco~Alejandro Lagunas-Rangel.
\newblock Neutrophil-to-lymphocyte ratio and lymphocyte-to-c-reactive protein
  ratio in patients with severe coronavirus disease 2019 (covid-19): a
  meta-analysis.
\newblock \emph{Journal of medical virology}, 92\penalty0 (10):\penalty0
  1733--1734, 2020.

\bibitem[Lengerich et~al.(2021{\natexlab{a}})Lengerich, Caruana, and
  Aphinayanaphongs]{lengerich2021data}
Benjamin~J Lengerich, Rich Caruana, and Yin Aphinayanaphongs.
\newblock Data-driven patterns in protective effects of ibuprofen and ketorolac
  on hospitalized covid-19 patients.
\newblock \emph{medRxiv}, 2021{\natexlab{a}}.

\bibitem[Lengerich et~al.(2021{\natexlab{b}})Lengerich, Caruana, Peysakhovich,
  Horwitz, and Aphinyanaphongs]{lengerich2021neutrophil}
Benjamin~J Lengerich, Rich Caruana, Alex Peysakhovich, Leora Horwitz, and Yin
  Aphinyanaphongs.
\newblock Neutrophil lymphocyte ratio as a predictor of glucocorticoid
  effectiveness in covid-19 treatment.
\newblock \emph{medRxiv}, 2021{\natexlab{b}}.

\bibitem[Lengerich et~al.(2022{\natexlab{a}})Lengerich, Caruana, Nunnally, and
  Kellis]{lengerich2022death}
Benjamin~J Lengerich, Rich Caruana, Mark~E Nunnally, and Manolis Kellis.
\newblock Death by round numbers and sharp thresholds: How to avoid dangerous
  ai ehr recommendations.
\newblock \emph{medRxiv}, 2022{\natexlab{a}}.

\bibitem[Lengerich et~al.(2022{\natexlab{b}})Lengerich, Nunnally,
  Aphinyanaphongs, Ellington, and Caruana]{lengerich2022automated}
Benjamin~J Lengerich, Mark~E Nunnally, Yin Aphinyanaphongs, Caleb Ellington,
  and Rich Caruana.
\newblock Automated interpretable discovery of heterogeneous treatment
  effectiveness: A covid-19 case study.
\newblock \emph{Journal of biomedical informatics}, 130:\penalty0 104086,
  2022{\natexlab{b}}.

\bibitem[Liu et~al.(2020)Liu, Du, Chen, Jin, Peng, Wang, Luo, Chen, and
  Zhao]{liu2020neutrophil}
Yuwei Liu, Xuebei Du, Jing Chen, Yalei Jin, Li~Peng, Harry~HX Wang, Mingqi Luo,
  Ling Chen, and Yan Zhao.
\newblock Neutrophil-to-lymphocyte ratio as an independent risk factor for
  mortality in hospitalized patients with covid-19.
\newblock \emph{Journal of Infection}, 81\penalty0 (1):\penalty0 e6--e12, 2020.

\bibitem[Mei et~al.(2021)Mei, van Wijk, Pham, and Marin]{mei2021role}
Zhen~W Mei, Xander~MR van Wijk, Huy~P Pham, and Maximo~J Marin.
\newblock Role of von willebrand factor in covid-19 associated coagulopathy.
\newblock \emph{The journal of applied laboratory medicine}, 2021.

\bibitem[Nori et~al.(2019)Nori, Jenkins, Koch, and
  Caruana]{nori2019interpretml}
Harsha Nori, Samuel Jenkins, Paul Koch, and Rich Caruana.
\newblock Interpretml: A unified framework for machine learning
  interpretability.
\newblock \emph{arXiv preprint arXiv:1909.09223}, 2019.

\bibitem[Orsi et~al.(2021)Orsi, Lijfering, Geersing, Rosendaal, Dekkers,
  le~Cessie, and Cannegieter]{orsi2021glucocorticoid}
Fernanda~A Orsi, Willem~M Lijfering, Geert-Jan Geersing, Frits~R Rosendaal,
  Olaf~M Dekkers, Saskia le~Cessie, and Suzanne~C Cannegieter.
\newblock Glucocorticoid use and risk of first and recurrent venous
  thromboembolism: self-controlled case-series and cohort study.
\newblock \emph{British Journal of Haematology}, 2021.

\bibitem[REMAP-CAP et~al.(2021)REMAP-CAP, ACTIV-4a, and
  Investigators]{remap2021therapeutic}
REMAP-CAP, ACTIV-4a, and ATTACC Investigators.
\newblock Therapeutic anticoagulation with heparin in critically ill patients
  with covid-19.
\newblock \emph{New England Journal of Medicine}, 385\penalty0 (9):\penalty0
  777--789, 2021.

\bibitem[Smith et~al.(2020)Smith, Trigonis, Rahman, Garcia, Salgado, Porter,
  Anderson, Duncan, Patel, House, et~al.]{smith2020bivalirudin}
N~Smith, R~Trigonis, O~Rahman, J~Garcia, J~Salgado, S~Porter, E~Anderson,
  M~Duncan, M~Patel, S~House, et~al.
\newblock Bivalirudin as therapeutic anticoagulation in covid-19 patients on
  ecmo.
\newblock \emph{ASAIO Journal}, pages 15--15, 2020.

\bibitem[Tang et~al.(2020)Tang, Li, Wang, and Sun]{tang2020abnormal}
Ning Tang, Dengju Li, Xiong Wang, and Ziyong Sun.
\newblock Abnormal coagulation parameters are associated with poor prognosis in
  patients with novel coronavirus pneumonia.
\newblock \emph{Journal of thrombosis and haemostasis}, 18\penalty0
  (4):\penalty0 844--847, 2020.

\end{thebibliography}

\clearpage
\appendix
\renewcommand{\thesubfigure}{\alph{subfigure}}

\setcounter{table}{0}
\setcounter{figure}{0}
\renewcommand{\thesection}{S\arabic{section}}  
\renewcommand{\thetable}{S\arabic{table}}  
\renewcommand{\thefigure}{S\arabic{figure}}

\onecolumn

\section{Dataset Construction}
\label{sec:s1}

\paragraph{Pre-Processing}
Our dataset consists of 11080 total hospitalized patients who have lab-confirmed cases of Covid-19. To filter out patients who were hospitalized for reasons other than Covid-19, we excluded patients who have indicators of (1) pregnancy: outpatient prenatal vitamins, in-patient oxytocics, folic acid preparations; or (2) scheduled surgery: urinary tract radiopaque diagnostics, laxatives, general anesthetics, antiemetic/antivertigo agents, or antiparasitics. We also require that the patients have recorded temperature, age, BMI, and Admission Day. Finally, we remove patients who died within six hours of admission.
The patient population changed over time. The majority of patients were admitted from days 30-50. As Figure~\ref{fig:cohort} shows, this period also contained the majority of patients with extremely high risk. 

\begin{figure}[htb]
\floatconts
  {fig:cohort}
  {\caption{Predicted probability of mortality and admission day for each patient. In this figure, each point represents an individual patient, with the vertical location indicating the probability of mortality predicted by the mortality risk model. There is a high density of patients in the first 60 days of the pandemic, and the most high-risk patients were also observed at that time.}}
  {\includegraphics[width=0.35\linewidth]{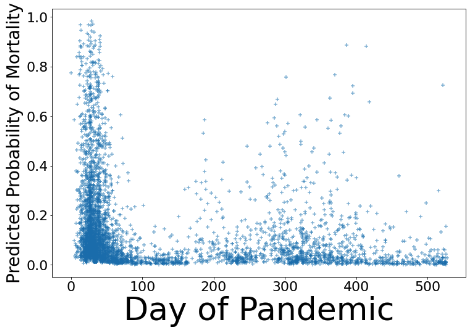}}
\end{figure}

To correct for patient risk confounding, we observe pre-admission features including demographics, comorbidities, outpatient medications, initial in-patient vitals, and initial in-patient lab tests. We exclude any measurement taken within 24 hours of the patient mortality. The 45 total features are listed below.

\begin{itemize}
    \item Demographics/Vitals:
    \begin{itemize}
        \item Age
        \item Sex
        \item BMI
        \item Day
        \item Temperature
    \end{itemize}
    \item Comorbidities:
    \begin{itemize}
        \item Myocardial Infarction
        \item Congestive Heart Failure
        \item Peripheral Vascular Disease
        \item Cerebrovascular Disease
        \item Dementia
        \item Chronic Obstructive Pulmonary Disease
        \item Peptic Ulcer Disease
        \item Mild Liver Disease
        \item Diabetes without chronic complications
        \item Diabetes with chronic complications
        \item Hemiplegia or paraplegia
        \item Renal disease
        \item Cancer (any malignancy)
        \item Metastatic solid tumor
        \item Charlson score
        \item Hypertension
        \item Atrial fibrillation
        \item Valve Replacement
        \item Rheumatoid Arthritis
    \end{itemize}
    \item Outpatient Medications taken before hospitalization (limited to medication classes taken by at least 100 patients):
    \begin{itemize}
        \item Antihyperglycemic, Biguanide Type
        \item Laxatives And Cathartics
        \item Platelet Aggregation Inhibitors
        \item Vitamin D Preparations
        \item Calcium Channel Blocking Agents
        \item Proton-Pump Inhibitors
        \item Antihyperlipidemic-Hmgcoa Reductase Inhib (Statins)
        \item Beta-Adrenergic Agents, Inhaled, Short Acting
        \item Blood Sugar Diagnostics
        \item Anticonvulsants
        \item Analgesic/Antipyretics,Non-Salicylate
        \item Beta-Adrenergic Blocking Agents
    \end{itemize}
    \item Lab Values:
    \begin{itemize}
        \item Potassium (0.5\% Missing)
        \item Ferritin (8.3\% Missing)
        \item Calcium (0.5\% Missing)
        \item Neutrophil \% (0.0\% Missing)
        \item Lymphocyte \% (0.0\% Missing)
    \end{itemize}
\end{itemize}


\clearpage
\section{Extended Description of Methods}

\begin{figure*}[htbp]
\floatconts
  {fig:arch}
  {\caption{Model Architecture. We use a generalized additive model to estimate homogeneous (static) risk factors, and an interaction with day (day of pandemic) to estimate time-varying risk factors. }}
  {\includegraphics[width=0.7\linewidth]{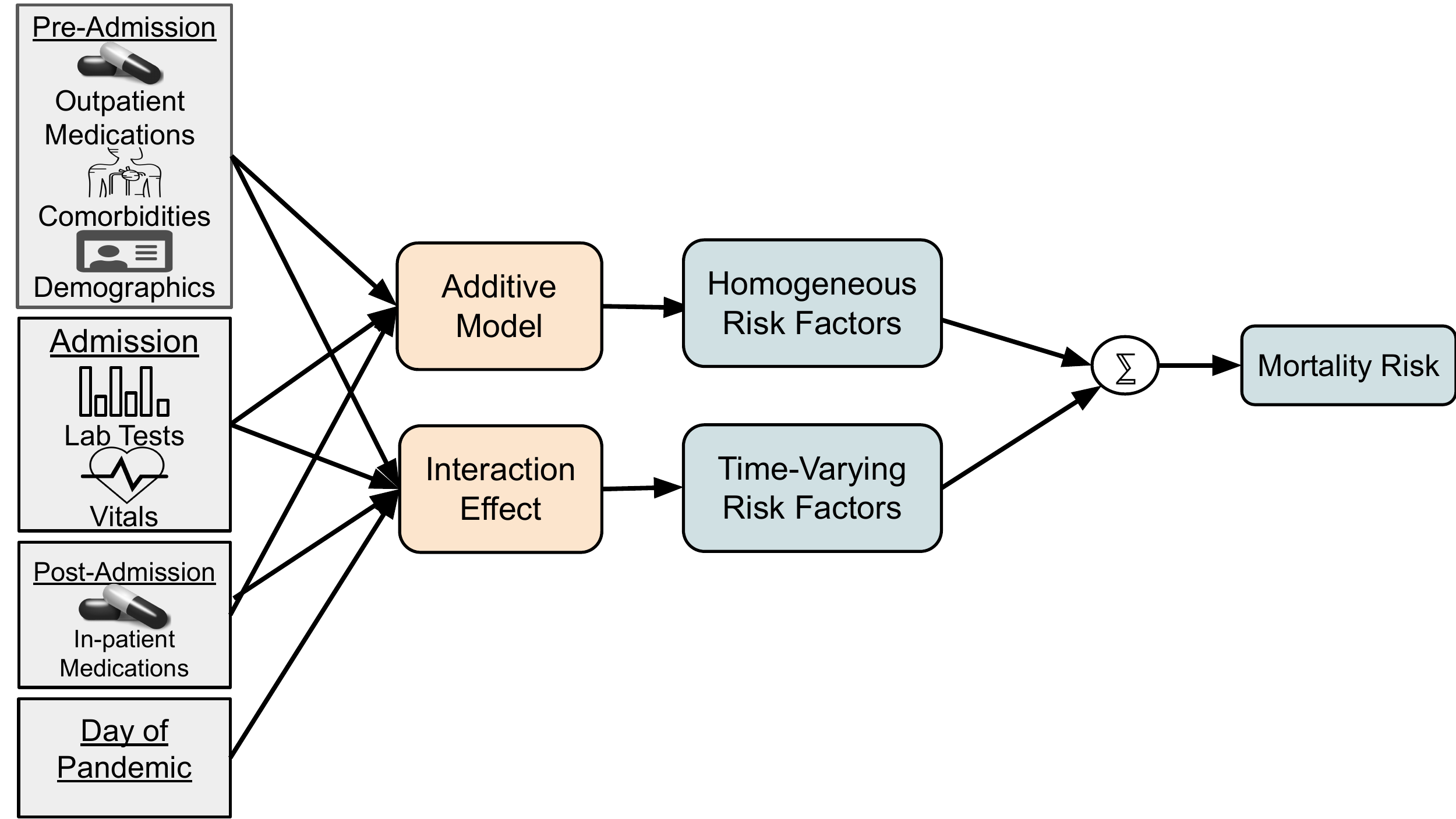}}
\end{figure*}

\begin{figure}[h]
\floatconts
  {fig:binarize}
  {\caption{Effects of continuous lab values before binarization to biomarker rules. In both panes, blue areas represent reduced risk of mortality while red areas represent increased risk of mortality. {\bf (A)} Low calcium is consistently associated with increased risk of mortality. {\bf (B)} Low hematocrit levels were originally associated with increased risk of mortality, but after 50 days, the association changed sharply and high hematocrit became associated with increased risk of mortality.}}
  {\includegraphics[width=\linewidth]{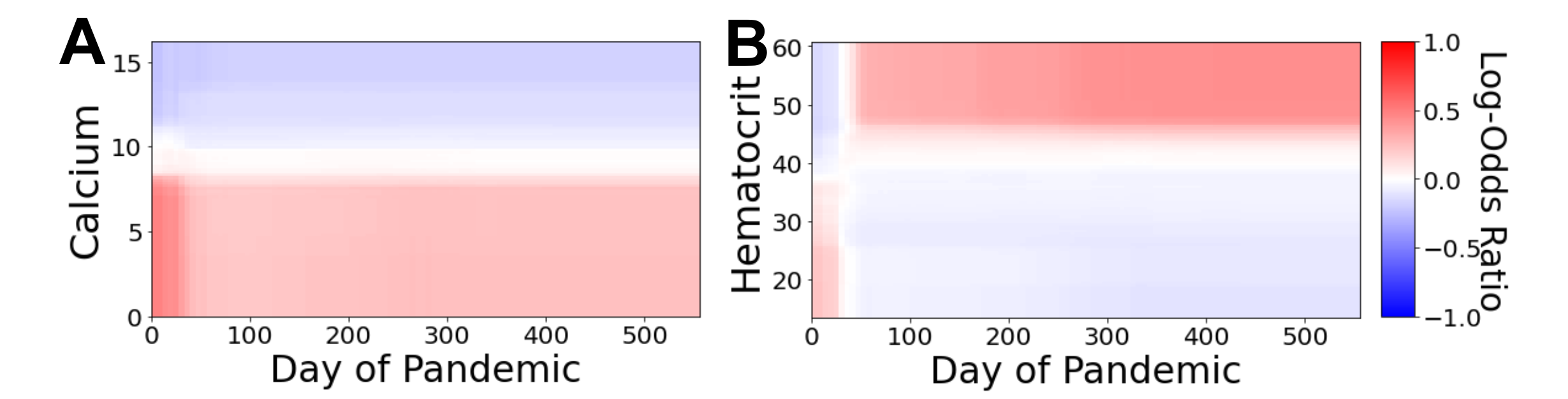}}
\end{figure}

The GAM provides odds ratios (ORs) for the mortality risk associated with each biomarker on each day. 
We supplement these daily ORs with ORs from 3 logistic regression (LR) models each trained on approximately one third of the patients (LR1 was trained on patients hospitalized from day 1 to day 100, LR2 was trained on patients hospitalized from day 100 to 300, while LR3 was trained on patients hospitalized from day 300 to 527) and observe qualitatively similar patterns in the GAM and the LR models, although the GAM provides more statistical power and better temporal resolution.

\clearpage
\section{Extended Results}

\begin{figure*}[ht]
    \floatconts
    {fig:in_hospital_meds}
    {\caption{Effects of In-Hospital Medications, ordered by mean effect size. In each pane, we plot the additive effect on mortality log-odds (lower is more protective). The only medications which appear consistently helpful are Ketorolac and Ibuprofen, although the effectiveness of these medications appear to decrease.}}
    {\centering
    \includegraphics[width=0.9\textwidth]{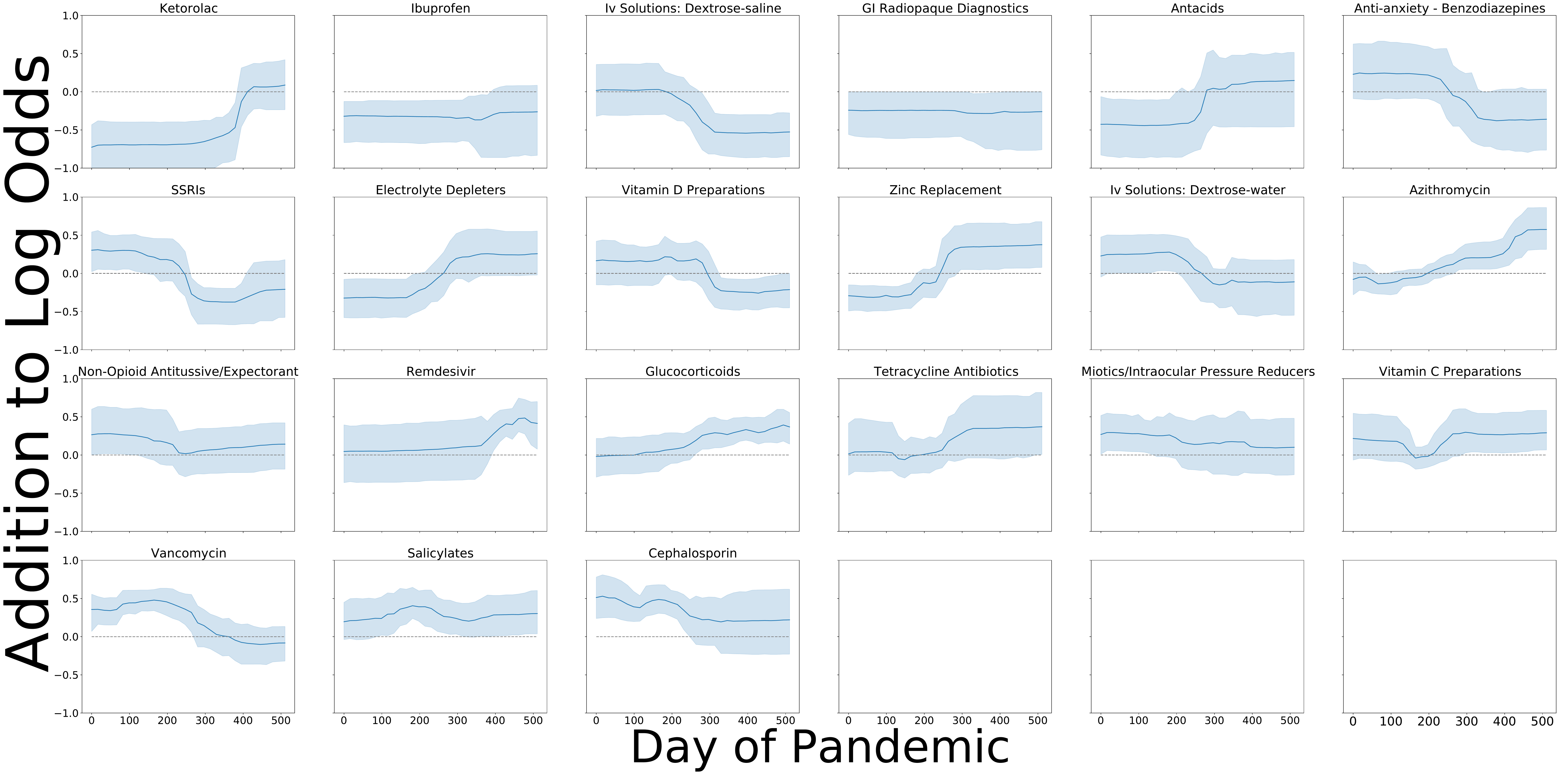}}
\end{figure*}

\begin{figure*}[ht]
    \floatconts
    {fig:preadmit}
    {\caption{Effects of Comorbidities and Pre-hospitalization medications.}}
    {\centering
    \subfigure[Comorbidities][b]{
        \includegraphics[width=0.8\textwidth]{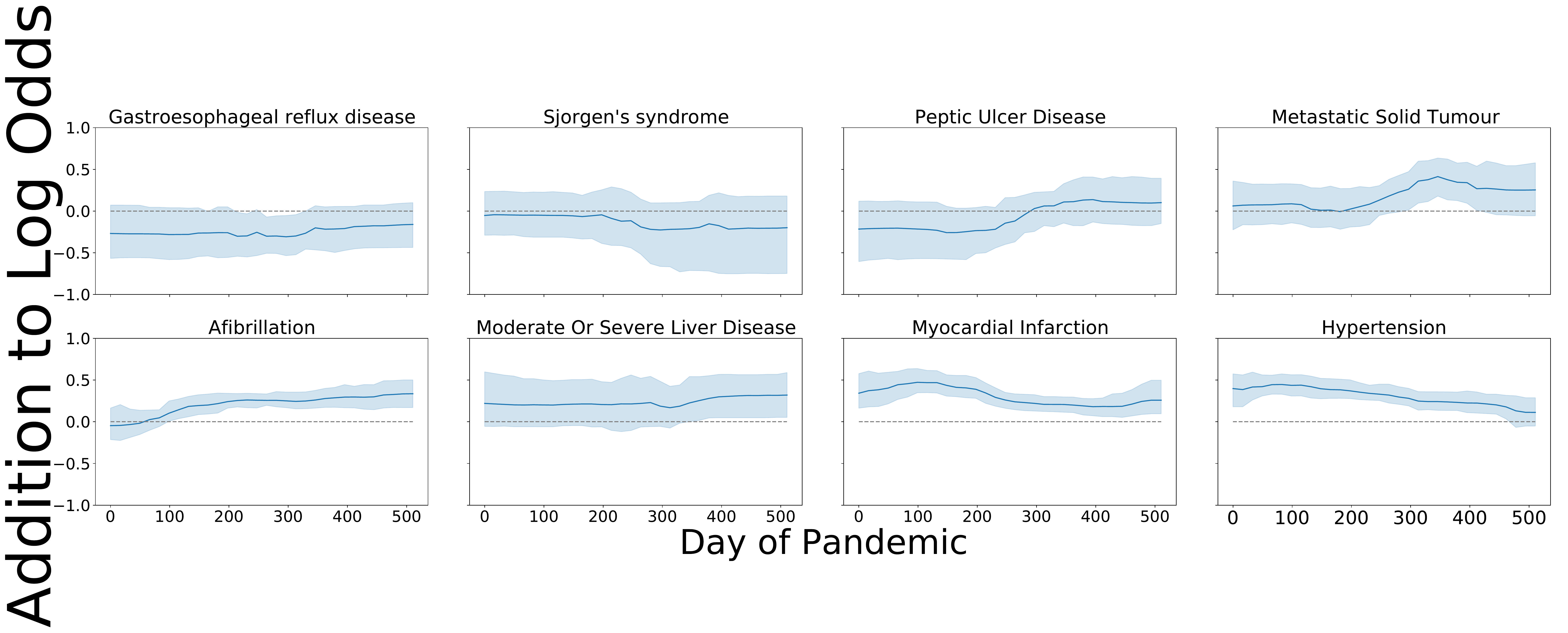}
    }
    ~
    \subfigure[Pre-hospitalization medications][b]{
        \includegraphics[width=0.8\textwidth]{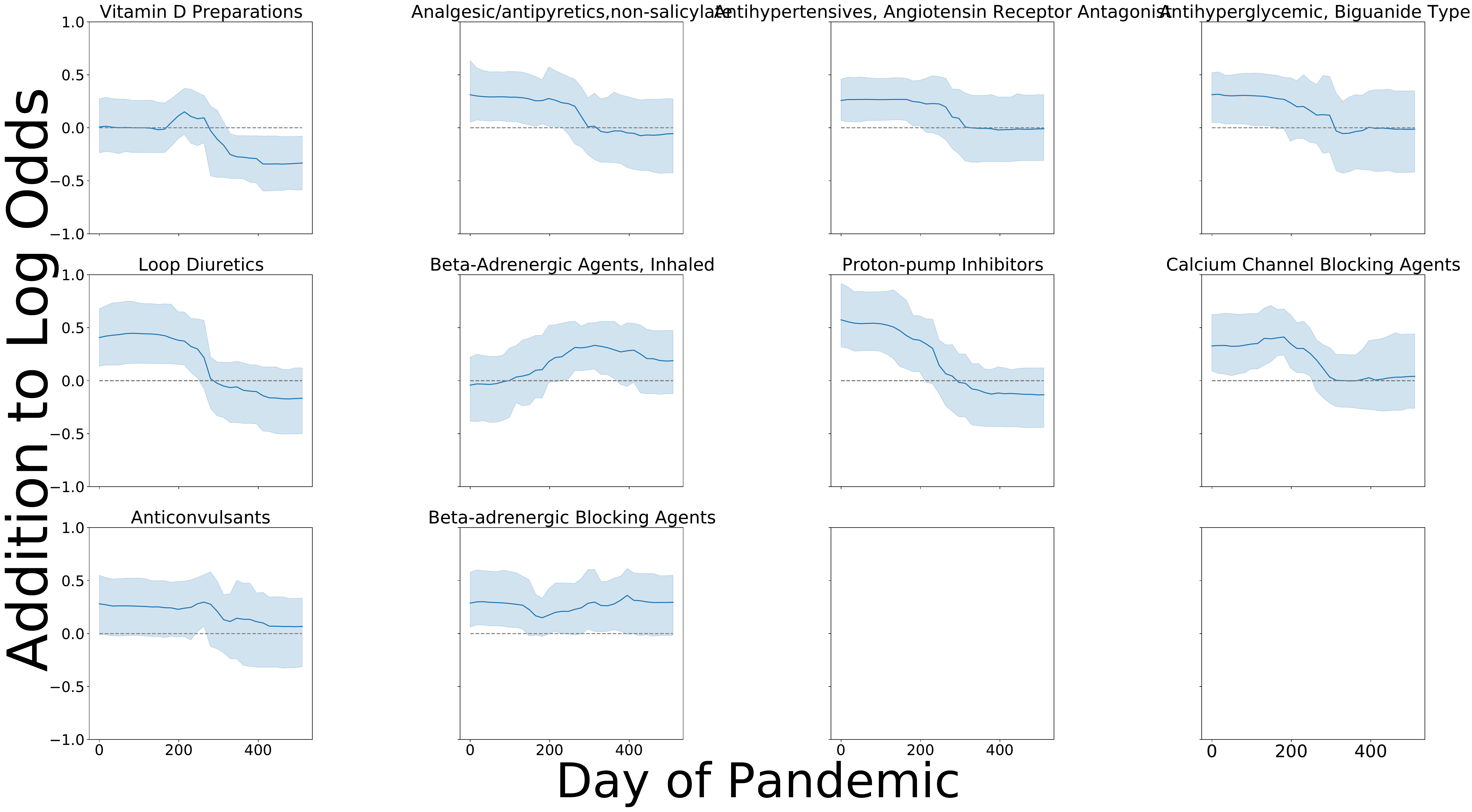}
    }}
\end{figure*}

\subsection{Decreasing Effectiveness of Anti-Inflammatory Medications}
Just as the predictive power of inflammation reduced over time, so too did the apparent effectiveness of anti-inflammatory medications (Figure~\ref{fig:in_hospital_meds}) reduce over time. 
In particular, glucocorticoids shifted from having no association with any significant homogeneous effect to an association with a deleterious homogeneous. 
This could correspond to glucocorticoids having a beneficial effect for only a subset of the patient population \citep{lengerich2021neutrophil}, but being used as standard care.
In addition non-steroidal antiflammatory drugs (NSAIDs) Ketorolac and Ibuprofen shifted from an association with decreased mortality to no significant association (confidence intervals that overlap zero effect). 
This corresponds to previously-observed benefits of NSAIDs \citep{lengerich2021data} being reduced by anti-inflammatory steroids becoming a standard of care.

\subsection{Comorbidities and Pre-Hospitalization Medications}
The baseline mortality risks of comorbidities and pre-admission out-patient medications are shown in Figure~\ref{fig:preadmit}. 
In general, these effects of comorbidities and out-patient medications concord with mechanisms of mortality involving inflammation and/or thromboses. 
The most protective effect is peptic ulcer disease, the effect of which is offset in some patients by a deleterious effect of proton-pump inhibitors. 
The second-most protective effect is platelet aggregation inhibitors (low-dose aspirin), which can prevent thromboses. 
The third-most protective effect is valve replacement, for which patients are continually on out-patient anti-coagulation medications. 
The most deleterious effects are dementia, congestive heart failure, beta-adrenergic agents, and myocardial infarction. 

\clearpage
\section{Homogeneous Effects}

\begin{figure*}[htb]
\floatconts
  {fig:homogeneous}
  {\caption{Homogeneous (static) effects of lab values and comorbidities. {\bf (A-P)} Effects of lab values, with shaded regions representing 95\% confidence intervals. Each pane displays the uncorrected effect estimated by univariable marginalization (gray) and by the multivariable GAM (red). Each yellow tick mark along the horizontal axis denotes 10 patients. {\bf (Q)} Estimated mortality odds ratio associated with comoribidities and outpatient (pre-hospitalization) medications, with 95\% confidence intervals.}}
  {\includegraphics[width=0.95\linewidth]{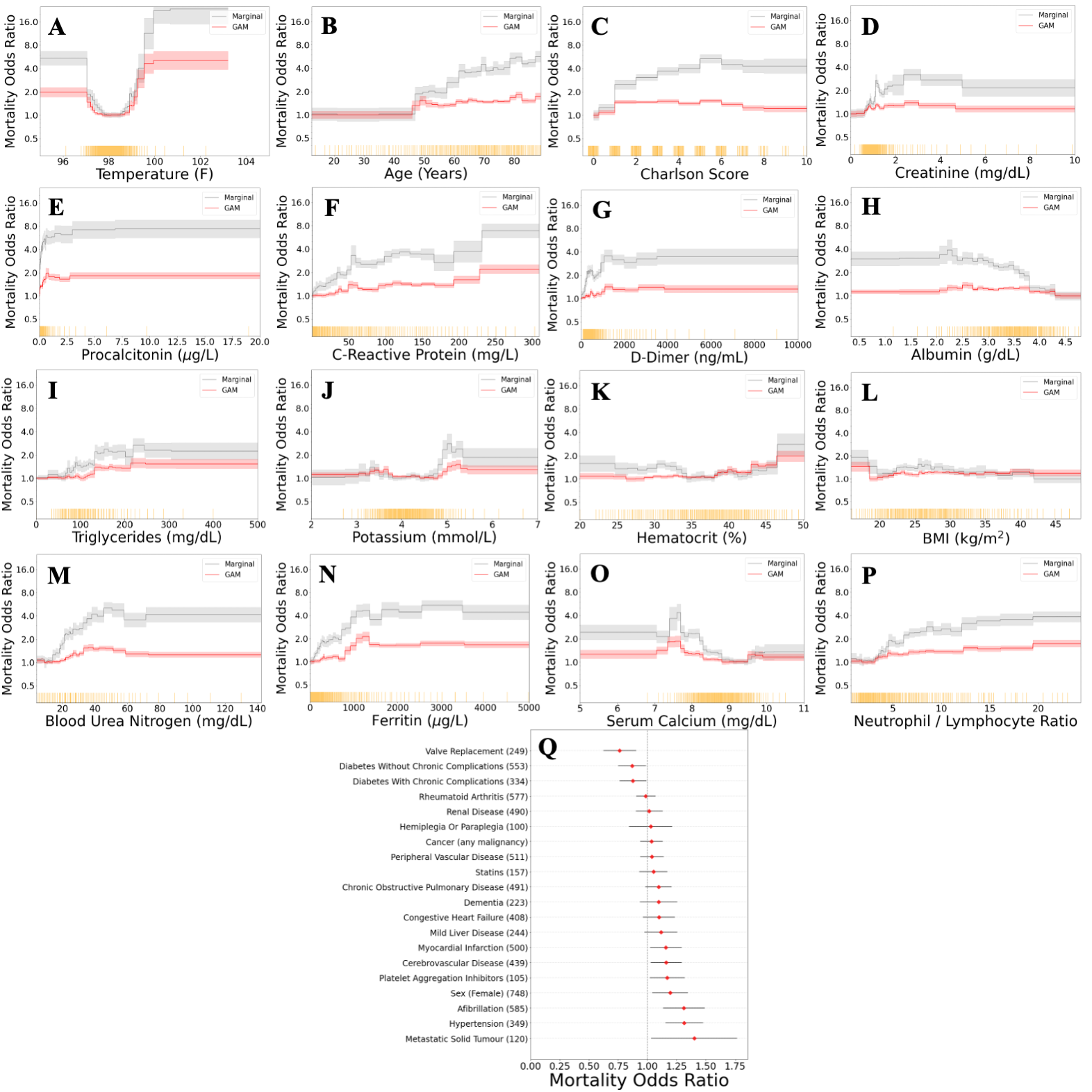}}
\end{figure*}

\end{document}